%% file: main4arxiv.tex
\documentclass[11pt]{article}

\newif\ifneuripsstyle
\neuripsstyletrue

\ifneuripsstyle
  \IfFileExists{neurips_2024.sty}{
    \usepackage[preprint]{neurips_2024}
  }{
    \usepackage[margin=1in]{geometry}
  }
\else
  \usepackage[margin=1in]{geometry}
\fi
\usepackage{amsfonts,amsbsy,amsmath,amssymb,amsthm}
\usepackage{bm}
\usepackage[nopatch=footnote]{microtype}
\usepackage{graphicx}
\usepackage{booktabs}
\usepackage{tabularx}
\usepackage{enumitem}
\usepackage{algorithm}
\usepackage{algorithmic}
\usepackage{natbib}
\usepackage[colorlinks=true,linkcolor=blue,citecolor=blue,urlcolor=blue]{hyperref}
\usepackage{lineno}
\usepackage{placeins}

\newif\ifreview
\reviewfalse

\usepackage{setspace}
\newtheorem{proposition}{Proposition}

\theoremstyle{definition}

\newtheorem{remark}{Remark}

\newcommand{\R}{\mathbb{R}}

\newcommand{\Var}{\mathbb{V}\mathrm{ar}}

\newcommand{\Prob}{\mathbb{P}}

\newcommand{\norm}[1]{\left\lVert #1 \right\rVert}
\newcommand{\abs}[1]{\left\lvert #1 \right\rvert}
\newcommand{\sgn}{\mathrm{sgn}}
\newcommand{\argmin}{\mathop{\mathrm{arg\,min}}}
\newcommand{\argmax}{\mathop{\mathrm{arg\,max}}}

\newcommand{\X}{\mathbf{X}}

\newcommand{\betaVec}{\bm{\beta}}
\newcommand{\keywords}[1]{\bigskip\noindent \emph{Keywords:} #1}

\numberwithin{equation}{section}

\title{Stagewise Adaptive Boosting}
\author{Valerie R. Fu, Brayden Lu, and Lu Tian}
\date{\today}

\begin{document}
\ifreview\linenumbers\fi
\maketitle

\begin{abstract}
Stagewise algorithms and sparsity-inducing regularization are two influential approaches to high-dimensional modeling. Existing stagewise theory, however, largely centers on convex regularization, whereas stronger sparsity and other structural preferences are often pursued through fixed-level nonconvex optimization rather than generating an entire adaptive solution path. We develop stagewise adaptive boosting, a general framework that builds a monotone path by selecting the update consistent with structural preference. The framework can encourage sparsity along the path itself and also admits extensions that promote other properties of the solution, including group selection, similarity across models for multiple outcomes, and known priorities or prerequisite relations among predictors. We study its theoretical property as well as finite sample performance using simulations and real-data analyses. Across these studies, the proposed pathwise approach often achieves predictive performance comparable to commonly used baselines while selecting substantially smaller models. These results suggest that adaptive stagewise learning provides a practical route to stronger sparsity and richer structural control without repeatedly solving separate weighted or nonconvex optimization problems.
\end{abstract}

\keywords{concave regularization; differential inclusion; early stopping; high-dimensional data; nonconvex optimization; regularization paths; structured sparsity; variable selection}

\section{Introduction}
Boosting and sparsity-inducing regularization are two central paradigms for
high-dimensional modeling.
In regression and classification, boosting builds an additive predictor by
repeatedly fitting weak learners to current residuals or negative gradients of
the loss \citet{friedman2001greedy,buhlmann2007boosting,buhlmann2011hd}.
Penalized likelihood methods such as the lasso
\citep{tibshirani1996lasso}, in contrast, induce sparsity through an explicit
constraint or penalty on the coefficients.
A major theme in the literature is that these viewpoints meet in stagewise
algorithms: for least-squares regression, \citet{efron2004lars} and
\citet{rosset2004boosting} showed that, under suitable regularity conditions, the forward stagewise path with infinitesimal step
size coincides with the lasso solution path, and related connections have been
developed for generalized linear models and other losses
\citep{zhao2007stagewise,tibshirani2015stagewise}.
As a result, conventional stagewise theory and algorithms largely inherit the
geometry of convex regularization, especially the $\ell_1$ penalty, and tend to produce sparse models.

This perspective leaves an important gap.
In many applications, one wants a single solution path that remains sparse as
it evolves and that can also encode structural priorities among predictors.
The lasso and related convex regularization are effective and computationally stable, but when tuned for prediction they can retain too many variables
\citep{tibshirani1996lasso,leng2006modelsel}.
Moreover, existing stagewise constructions are primarily designed for convex
budget sets, so they do not directly provide a general pathwise mechanism encouraging stronger sparsity or richer structural preferences.

Concave regularization offers one route to stronger sparsity.
Existing work on nonconvex penalties, including the bridge penalty
\citep{frank1993chemometrics,fu1998bridge}, SCAD \citep{fanli2001scad}, and
MCP \citep{zhang2010mcp}, shows that relaxing convex geometry can reduce
shrinkage on large coefficients and yield smaller, less biased models.
But this literature has mostly been developed through special-case penalties
and fixed-level nonconvex optimization, often via local linear approximation or
specialized coordinate descent
\citep{zou2008onestep,breheny2011coorddesc,mazumder2011sparsenet}.
What is still missing is a general adaptive stagewise framework that produces a
single monotone path while encouraging sparsity throughout the path itself.

A related idea appears in adaptive lasso and adaptive boosting methods, which
use data-driven weights to prioritize predictors.
Those methods are valuable, but they are typically implemented as a two-step
procedure: first obtaining pilot information and then solving a second,
weighted optimization problem. If the priority information from the first step is not accurate, then the performance of the final prediction model may be poor.  
Our goal is different. We seek one pathwise procedure that adaptively reallocates budget as the model updates, so that predictor priorities and sparsity are encouraged continuously
rather than imposed after a preliminary fit.

This paper develops such a flexible framework, which we call \emph{stagewise adaptive
boosting} (SAB).
The base algorithm builds a single monotone solution path adaptively updated favoring active directions that deliver the largest
first-order loss decrease. The SAB algorithm is sufficiently flexible to encourage a variety of structural preferences. Classical bridge-type regularizations arise as a special example.

\paragraph{Contributions.}
Our first contribution is a general adaptive stagewise algorithm for promoting
sparsity along a single solution path.
We propose coordinate update according to an adaptive gradient score that
encourages model sparsity. Our second contribution is to show that the same framework extends naturally to encourage other structural properties beyond plain sparsity. In particular, we develop extensions that favor group selection, similarity across models for multiple outcomes, and known priorities or prerequisite relations among predictors, such as strong heredity for interactions. 

The remainder of the paper reviews related work (Section~\ref{sec:background}),
develops the bridge-functional viewpoint, update rule, and algorithm
(Section~\ref{sec:formulation}),
records theoretical properties (Section~\ref{sec:theory}),
develops structured extensions (Section~\ref{sec:extensions}),
and evaluates empirical trade-offs (Sections~\ref{sec:sim}--\ref{sec:data}).
We conclude with a discussion in Section~\ref{sec:discussion}.
The Appendix collects additional derivations, extension-specific
simulations and ablations, and extended empirical results.

\section{Background and Related Work}\label{sec:background}
This section briefly reviews penalized M-estimation, concave penalties with
$\ell_\gamma$-type geometry, and stagewise boosting algorithms.

\paragraph{Set-up and M-estimation}
Let $(x_i,y_i)_{i=1}^n$ be independent observations with predictors
$x_i\in\R^p$ and responses $y_i\in\R$.
We consider $f_\beta(x)$, a class of models indexed by a coefficient vector $\betaVec\in\R^p,$ and a
convex, twice continuously differentiable loss function
$\ell(y, f_\beta(x))$.
For concreteness one may think of linear predictors $f_\beta(x)=x^\top\beta$
with squared-error or logistic loss, but the framework extends to general
smooth convex losses. The M-estimation estimates $\beta$ or $f_\beta(x)$ by minimizing the empirical loss:
\[
  L(\betaVec) = \frac{1}{n}\sum_{i=1}^n
  \ell\bigl\{y_i, f_\beta(x_i)\bigr\}.
\]
\paragraph{Regularization}
When the dimension of $x$ is not small, penalized M-estimation is often used to minimize a regualrized loss 
\[
  \hat\beta(\lambda)
  \in \argmin_{\betaVec\in\R^p}
  \Bigl\{
    L(\betaVec) + \lambda P(\betaVec)
  \Bigr\},
\]\relax
where $P(\betaVec)$ is a penalty functional, and $\lambda\ge 0$ is the tuning
parameter controlling the strength of regularization.
 The lasso penalty $P(\betaVec)=\norm{\betaVec}_1$ is a special case with fast optimization algorithms and desirable statistical properties.
\citep[e.g.,][]{tibshirani1996lasso,buhlmann2011hd}. An equivalent constrained formulation replaces the penalty with a budget
constraint
\[
  \hat\beta(B_0)
  \in \argmin_{\betaVec\in\R^p}
  \Bigl\{
    L(\betaVec) : P(\betaVec)\le B_0
  \Bigr\},
\]\relax
and for convex $P(\cdot)$ there is a one-to-one correspondence between
$\lambda$ and $B_0$ under mild regularity conditions.
Stagewise algorithms can be naturally interpreted as schemes that move along
such constrained solution paths by gradually increasing the budget cap from
zero.

\paragraph{Concave penalties}
Concave penalties modify the geometry of the constrained problem by penalizing large coefficients less than small ones. 
For example, the bridge functional 
$$ P_\gamma(\betaVec)=\sum_{j=1}^p |\beta_j|^\gamma  $$
is a concave penalty when $0<\gamma<1.$ 
In the corresponding constrained formulation, the level sets
$\{\betaVec: P_\gamma(\betaVec)\le B_0\}$ for $\gamma<1$ are star-shaped regions that are more sharply peaked around the coordinate axes than the $\ell_1$-diamond.
This induces a stronger preference for sparse configurations and tends to
concentrate the available budget on a few large coefficients.
This paper develops a stagewise algorithm that traces a path indexed by the
bridge-functional value $B(\betaVec)=P_\gamma(\betaVec)$. 

Concave penalties can reduce bias for large coefficients and, under suitable
sparsity assumptions, can achieve selection consistency and oracle properties
\citep[e.g.,][]{fanli2001scad,zhang2010mcp}.
However, they lead to nonconvex optimization problems, and standard convex
duality arguments no longer apply.
From a computational perspective, popular approaches include local linear
approximation (iterative reweighted $\ell_1$) schemes and coordinate descent
algorithms tailored to specific nonconvex penalties
\citep{zou2008onestep,breheny2011coorddesc,mazumder2011sparsenet}.
These methods target solutions at fixed penalty levels, whereas our focus is on constructing a solution path.

\paragraph{Stagewise boosting algorithms.}
Forward stagewise regression builds an additive predictor by repeatedly
updating one coordinate in the direction of steepest loss descent.
For squared-error loss with infinitesimal steps, this path coincides with the
lasso path under suitable conditions
\citep{efron2004lars,rosset2004boosting}.
More generally, gradient boosting fits a base learner $h(\cdot;\theta)$ to the
negative gradient at the current predictor and updates by adding a small
multiple of that learner \citep{friedman2001greedy,buhlmann2007boosting}.
With coordinate-function learners $\{x\mapsto x_j\}_{j=1}^p$, it reduces to a
coordinate-wise stagewise procedure.

Several authors study links between stagewise algorithms and
$\ell_1$-type regularization
\citep{rosset2004boosting,zhao2007stagewise,tibshirani2015stagewise}.
Under a monotonicity condition that rules out sign changes, forward stagewise
trajectories can approximate the path of an $\ell_1$-constrained problem.
This view organizes stagewise methods by cumulative regularization along the
path. A more related perspective is given by \citet{friedman2012fast} on computationally simple stagewise search rules for tracing regularization paths under a broad class of loss functions and penalties. Our formulation is similar in spirit, but we make the coordinate-priority rule explicit via a general weight function $p(\cdot)$ including the regularization framework as a special case.

Much less is known about stagewise procedures for concave penalties.
Existing nonconvex methods usually target stationary points at fixed
regularization levels via local linear approximation or specialized
coordinate-descent schemes
\citep{zou2008onestep,fanli2001scad,zhang2010mcp,breheny2011coorddesc,mazumder2011sparsenet}.

By contrast, our aim is to construct a general adaptive stagewise path that
need not correspond exactly to a convex or even concave regularization, but
still encourages sparse solutions by giving higher updating priority to active
coordinates. We further develop extensions that encourage additional
structural properties along the solution path.


\section{Method}\label{sec:formulation}
We now formalize the problem setup and introduce notation.
We work with the same setup as in the previous section,
and we write
$g(\betaVec)=\nabla L(\betaVec)\in\R^p$ for its gradient, with components
$g_j(\betaVec)=\partial L(\betaVec)/\partial\beta_j$.

\subsection{Base Algorithm}
Our starting point is a general adaptive stagewise scheme.  To this end, let
$p(\cdot):\R^+\to[\delta_0,\infty)$ be a monotone increasing function, where
$\delta_0>0$ is a baseline step size. At a current coefficient vector
$\betaVec$, we consider one-coordinate updates of the form
\[
  \tilde\betaVec=\betaVec+s_j\,p(\abs{\beta_j})e_j,
\]
where $s_j\in\{-1,1\}$ is chosen to reduce the loss and to preserve the sign
of an already active coefficient and $e_j$ is a zero/one vector indicating the $j$th coordinate. This family yields a general stagewise
procedure that favors reuse of the active set, since
larger active coefficients receive larger admissible one-step moves. To select a coordinate to update, consider  the first-order approximation to the loss decrease corresponding to the update described above: 
\[
  -\Delta L_j
  \approx
  \abs{g_j(\betaVec)}\,\Delta\beta_j
  = \abs{g_j(\betaVec)}\,p(\abs{\beta_j}).
\]\relax
 Therefore, an adaptive stagewise score for coordinate selection can be defined as
\[
  S_j(\betaVec)
  =
  \begin{cases}
      \abs{g_j(\betaVec)}\,p(\abs{\beta_j}),
      & \beta_j\neq 0
        \text{ and }
        \sgn(\beta_j)=-\sgn(g_j(\betaVec))\\
    \abs{g_j(\betaVec)}\,p(0),
      & \beta_j=0\\
    -\infty,  & \beta_j\neq 0
        \text{ and }
        \sgn(\beta_j)=\sgn(g_j(\betaVec))
  \end{cases}.
\]\relax
for both active and inactive coordinates. The SAB algorithm selects, at each iteration, the
coordinate $j^\star$ that maximizes $S_j(\betaVec)$ and then updates
$\beta_{j^\star} \to \beta_{j^\star}+s_{j^\star}\Delta\beta_{j^\star}.$ 
When $p(\cdot)$ is constant, active and inactive coordinates are both ranked by
$\abs{g_j(\betaVec)}$, in line with classical forward stagewise regression.

The resulting base algorithm is summarized next.
\begin{algorithm}[t]
\caption{General SAB algorithm with monotone weight function $p(\cdot)$}
\label{alg:sab}
\begin{algorithmic}[1]
\STATE \textbf{Input:} data $(x_i,y_i)_{i=1}^n$, loss $\ell$, monotone
  increasing function $p:\R^+\to[\delta_0,\infty)$, baseline step size
  $\delta_0>0$, and stopping iteration $K$.
\STATE Initialize $\betaVec^{(0)}=0$ and $k=0$.
\WHILE{$k<K$}
  \STATE Compute gradient $g(\betaVec^{(k)})=\nabla L(\betaVec^{(k)})$.
  \FOR{$j=1,\dots,p$}
    \STATE Set the candidate step $\Delta\beta_j=p(\abs{\beta_j^{(k)}})$.
    \STATE Evaluate the adaptive stagewise score
      $S_j(\betaVec^{(k)})$.
  \ENDFOR
  \STATE Select $j^\star\in\argmax_j S_j(\betaVec^{(k)})$.
  \STATE Update $\betaVec^{(k+1)}=\betaVec^{(k)}+s_{j^\star}\Delta\beta_{j^\star}e_{j^\star}$,
    where $s_{j^\star}$ is the sign-preserving descent direction for coordinate $j^\star$.
  \STATE Set $k\leftarrow k+1$.
\ENDWHILE
\STATE \textbf{Output:} sequence of coefficient vectors $\{\betaVec^{(k)}\}$.
\end{algorithmic}
\end{algorithm}

In algorithm \ref{alg:sab}, we enforce a monotone update rule that disallows sign flips of active coefficients, consistent with standard forward-stagewise algorithm \citep{rosset2004boosting,zhao2007stagewise}. The same principles extend beyond linear coordinate dictionaries to richer base learners (e.g., splines or trees).
In practice, SAB can be stabilized and accelerated by a few standard refinements: use a learning rate $\nu\in(0,1]$ (replace $\Delta\beta_j$ by $\nu\Delta\beta_j$), and optionally subsample observations or coordinates. Per iteration, scoring all coordinates costs at least $O(p)$ plus gradient computation; for smooth losses this is typically $O(np)$, while for squared-error loss a precomputed Gram matrix $n^{-1}\X^\top\X$ enables incremental updates with $O(n+p)$ per step after a one-time $O(np^2)$ setup. 

\subsection{Implementation}\label{sec:budget-step}
Implementing SAB requires a specification of weight function $p(\cdot).$  An important choice can be motivated from the bridge-constrained problem
\begin{equation}\label{eq:constrained-problem}
  \hat\beta(B_0)
  = \argmin_{\betaVec\in\R^p}
  \Bigl\{
    L(\betaVec) : P_\gamma(\betaVec)\le B
  \Bigr\}, 
\end{equation}\relax
whose solution path as a function of $B$ is of interest. For $\gamma=1$ and squared-error loss, problem \eqref{eq:constrained-problem} reduces to $\ell_1$-constrained least squares, whose solution path can be traced exactly by LARS \citep{efron2004lars}. For $0<\gamma<1$ the feasible set is nonconvex, so exact solution paths may be discontinuous and difficult to compute. 
Nevertheless, \eqref{eq:constrained-problem} remains meaningful because
$B(\betaVec)$ provides an interpretable measure of model complexity. This
motivates choosing the coordinate update so that a one-coordinate move
increases $B(\betaVec)$ by exactly $\delta_0$. Thus, if $\tilde{\betaVec}$
satisfies $B(\tilde{\betaVec})-B(\betaVec)=\delta_0$, then
\begin{equation}\label{eq:coord-step-exact}
  p(|\beta_j|)
  = \Bigl(\abs{\beta_j}^\gamma+\delta_0\Bigr)^{1/\gamma}-|\beta_j|.
\end{equation}\relax

Equation~\eqref{eq:coord-step-exact} shows
that the induced step size depends on both the current coefficient magnitude
and $\gamma$. For $\gamma<1$, $p(|\beta_j|)$ is increasing in $|\beta_j|$, so
the same bridge-functional increment produces a larger change on an already
active coordinate than on a newly activated one.

Other monotone choices of $p(\cdot)$ are also possible, depending on how
aggressively the algorithm should favor active coordinates. Examples include
\begin{enumerate}
\item $p(|\beta|)=\delta_0 \log(1+\gamma|\beta|)$, 
\item $p(|\beta|)=\delta_0(1+\gamma_1|\beta|^{\gamma_2})$, and 
\item $p(|\beta|)=\delta_0 \exp(\gamma|\beta|)$.
\end{enumerate}

Now, we briefly discuss several practical aspects of implementing SAB with $p(\cdot)$ defined in (\ref{eq:coord-step-exact}).  The same principle applies to other family of functions for $p(\cdot).$  First, the concavity parameter $\gamma$ controls the
aggressiveness of the stagewise updates. Values close to $1$ behave similarly to $\ell_1$ regularization, while smaller values yield more concave penalties with stronger sparsity-inducing effects. In practice, it is natural to treat $\gamma$ as a tuning parameter to be selected by cross-validation or an information criterion. Second, the maximum length of the solution path should be chosen to cover the regime where the predictive performance stabilizes.
One can continuously monitor validation error and stop updating once overfitting becomes apparent.  Last, the bridge-functional increment $\delta_0$ controls the resolution of the stagewise path. Smaller $\delta_0$ more closely approximates the continuous-path limit but increases the number of iterations required to reach a given bridge-functional level. In practice we recommend choosing $\delta_0$ relative to the standardized feature scale, for example $\delta_0\in[10^{-3},\times 10^{-2}]\tilde{B}/p(\tilde{B})$, where $\tilde{B}$ is an upper bound of $\beta_j.$


\subsection{Theoretical Properties}\label{sec:theory}
Our theoretical focus is the sparsity gap between SAB updates
with an increasing weight function $p(\cdot)$ and the classical forward
stagewise rule, which corresponds to $p(\cdot)$ being constant.

Let $\betaVec_p(t)$ denote the sign-constrained stagewise path generated by a
monotone increasing function $p(\cdot)$, and let
$A_p(t)=\{j:\beta_{p,j}(t)\neq 0\}$ be its active set. Let
$\betaVec_c(t)$ denote the corresponding path under a constant weight function,
that is, the classical forward stagewise benchmark. We use the same
initialization for both paths and consider the infinitesimal-step limit. 
Suppose that up to time $t_0$ no active coordinate changes sign.

\begin{proposition}[Local expansion after switching to a constant weight]
\label{prop:local-switch}
Fix a point $\betaVec_p(t)$ on the SAB path and compare the
next local update chosen by the $p$-rule with the update that would be chosen
if, at that same point, one were to switch to a constant-weight rule. Then
switching from an increasing weight function $p(\cdot)$ to a constant weight
can only make expansion of the active set more likely at the next local step.
In particular, a coordinate that remains inactive under the constant-weight
rule would also remain inactive under the adaptive-rule, while some coordinates
that are not yet competitive under the adaptive-rule may become competitive after
the switch.
\end{proposition}

This is a consequence of the score comparison: an increasing weight function
$p(\cdot)$ gives already active coordinates an extra advantage through the
factor $p(|\beta_{p,j}(t)|)$ in the active score, relative to the constant-weight
rule:
$$ \frac{|S_j(\betaVec(t))|}{|S_i(\betaVec(t))|}\ge \frac{|g_j(\betaVec(t))|}{|g_i(\betaVec(t))|}$$
for features $i$ and $j$ from inactive and active sets, respectively. Therefore, increasing $p(\cdot)$ tends to delay new
variable entry, at least locally between successive switching events,
relative to the classical forward stagewise benchmark. This principle
explains why adaptive SAB often follows a sparser solution path, although we do
not claim a general theorem of support inclusion.

\begin{remark}
 A stronger global sparsity statement may also hold: if along each one-coordinate segment the relative
advantage of the active weighted score over every inactive score is preserved
in the sense that
\begin{equation}
  t\mapsto
  \frac{|g_i(\betaVec_p(t))|}
       {|g_j(\betaVec_p(t))|\,p(|\beta_{p,j}(t)|)} \label{eq:cond_crossmono}
\end{equation}
is nonincreasing until the next switching event, then the local delayed-entry
comparison propagates from segment to segment. Under such a segmentwise
cross-monotonicity condition, it is plausible that the entire SAB path remains
at least as sparse as the path obtained by switching to a constant weight, and
hence as sparse as the conventional stagewise path globally.  
\end{remark}

\subsection{Structured and Multi-Outcome Extensions of Stagewise Adaptive Boosting}
\label{sec:extensions}

This section extends the baseline stagewise adaptive boosting (SAB) path to
three common forms of domain structure: grouped predictors, multiple outcomes,
and ordered or hierarchical terms. In all three cases, we consider the the adpative coordinate update
from Section~\ref{sec:budget-step} as the basis. Additional structure consideration enters only through the competition rule. The principle is to reward candidate to form a desirable activation set. In the following, we will discuss the extension of base SAB algorithm in three different settings.

\paragraph{Extension I: Group-Aware Stagewise Adaptive Boosting}
For grouped predictors, let $\mathcal{G}=\{G_1,\dots,G_M\}$ denote the
collection of groups. The goal is to encourage features within the same group
to be selected or deselected together. Define the
bridge mass  assigned to
group $m$ by
\begin{equation}
\label{eq:group-share}
  \omega_m^{(k)}
  =   \frac{ \sum_{\ell\in G_m} p(|\beta_\ell^{(k)}|)}{\sum_{\ell=1}^p p(|\beta_\ell^{(k)}|)}.
\end{equation}\relax
Let $b_j^{(k)}$ be the baseline SAB score for $\beta_j$ in iteration $k.$ The new group-aware score is
\begin{equation}
\label{eq:group-score}
  S_j^{\mathrm{grp}}\bigl(\beta^{(k)}\bigr)
  =
  b_j^{(k)}\left\{1+\eta_g \sum_{m=1}^M \omega_{m}^{(k)}I(j\in G_m)\right\},
\end{equation}\relax
with $\eta_g\ge 0$. The selected coordinate at the $k$th iteration is therefore
$$
  j_k^\star \in \argmax_{1\le j\le p}
  S_j^{\mathrm{grp}}\bigl(\beta^{(k)}\bigr).
$$
Once $j_k^\star$ is chosen, the update is the same exact SAB step as in the
base method. Thus a coordinate is promoted when it belongs to a group that
already carries a large \emph{share} of the active bridge mass. Setting $\eta_{g}=0$ recovers baseline SAB exactly.

\paragraph{Extension II: Multi-Outcome Coupled SAB}
In the presence of related multiple outcomes, the intention is to encourage constructing models using similar sets of prediction features. To this end, 
 we use $q$ to index the outcome of interest and write $\bm{B}=[\beta^{(1)},\dots,\beta^{(Q)}]\in\R^{p\times Q}.$
Let $\beta_j^{(q, k)}$ and $b_{j,q}^{(k)}$ be the
$j$th coefficient for $q$th outcome at the $k$th iteration and its SAB score, respectively. Define
\begin{equation}
\label{eq:row-share}
  \omega_j^{(q, k)}
  =
    \sum_{r\neq q} \frac{p(|\beta_j^{(r, k)}|) }{\sum_{\ell=1}^p p(|\beta_\ell^{(r, k)}|)}.   
\end{equation}\relax
The modified score is
\begin{equation}
\label{eq:multi-score}
  S_{j,q}^{\mathrm{mt}}
  =
  b_{j,q}^{(k)}\left(1+\eta_{m}\omega_j^{(q, k)}\right),
\end{equation}\relax
where $\eta_m\ge 0$ controls the strength of cross-model
borrowing. If predictor $j$ is already active for another outcome, then
$w_j^{(q, k)}>0$ and the feature $j$ is promoted for outcome $q$. Cross-validation or other stopping criteria, such as BIC, may suggest different stopping for different outcomes. 

\paragraph{Extension III: Ordered and Hierarchical Selection Constraints}
We now incorporate prerequisite structure into the SAB path. This setting
includes ordered panels, hierarchical bases, and strong heredity for
interactions. Let $\mathcal{H}=(V,\mathcal{E}_H)$ be a directed acyclic graph
(DAG) on $V=\{1,\dots,p\}$, where an edge $i\to j$ means that term $j$
requires $i$. Let $\mathrm{pa}(j)=\{i\in V:(i\to j)\in\mathcal{E}_H\}$. Note that $\mathrm{pa}(j)$ can be empty set for some features.  The restriction for model building in this setting is that 
$$ i \in A \Rightarrow \mathrm{pa}(i)\subseteq A,$$
where $A$ represents the active set in the final model. To construct such a model, define
$A^{(k)}=\{j\in V:|\beta_j^{(k)}|>0\}$ at iteration $k.$ 
To enforce this rule algorithmically, restrict updates to the feasible set
\begin{equation}
\label{eq:order-closure}
  \mathcal{F}^{(k)}
  =
  \{j\in V:\mathrm{pa}(j)\subseteq A^{(k)}\},
\end{equation}\relax
 i.e., 
 $$
  j_k^\star
  \in
  \argmax_{j\in\mathcal{F}^{(k)}}
  b_j^{(k)}.
$$
Hence,the hard parent-first constraint is therefore exactly
enforced by restricting update to the candidate set.

\begin{remark}
A soft hierarchy variant keeps all coordinates eligible but rewards terms whose
parents are already active. Define the parent-activity share
\begin{equation}
\label{eq:order-share}
  \omega_j^{(k)}
  =
  \frac{\sum_{\ell\in \mathrm{pa}(j)} p(|\beta_\ell^{(k)}|)}
        {\sum_{\ell=1}^p p(|\beta_\ell^{(k)}|)},
\end{equation}\relax
and score by
\begin{equation}
\label{eq:order-score}
  S_j^{o}\bigl(\beta^{(k)}\bigr)
  = b_j^{(k)} \left\{(1-\eta_{o})+\eta_{o}\omega_j^{(k)}\right\},
\end{equation}\relax
where $\eta_{o}\in[0,1]$ controls softness: $\eta_{o}=0$
recovers unconstrained SAB scoring, and larger values increasingly favor
coordinates whose parents already carry nontrivial weights in the current model.
\end{remark}

The construction is transparent in the chain
$X_4\rightarrow X_6,  X_5\rightarrow X_6$. Under the hard rule, $X_6\notin
\mathcal{F}^{(k)}$ until both $X_4$ and $X_5$ are active, so $X_6$ cannot enter
early. Under the soft rule, $X_6$ can still compete but receives a larger bonus
as more of its parents become active. This setting is useful for building a regression model including interaction terms, which are discouraged unless the relevant main effects are included.


\paragraph{Tuning and implementation}
All structured versions are tuned by validation along the path indexed by the
bridge-functional level. In practice, this means selecting $\gamma$,
$\delta_0$, and any active coupling parameters
$(\eta_{g},\eta_{m},\eta_{o})$ while always
including the unstructured benchmark
$(\eta_{g},\eta_{m},\eta_{o})=(0,0,0)$. When
prerequisite constraints are used, it is also useful to compare against the
same fit with the prerequisite graph removed, so that the predictive cost of
the ordering restriction is visible.

\section{Simulation Studies}\label{sec:sim}
We report simulation experiments assessing the finite-sample predictive
behavior of SAB and the induced accuracy--sparsity trade-off as the concavity parameter $\gamma$ varies.

\subsection{Simulation study for the base algorithm}

\paragraph{Sparse linear regression.}
For each replicate we generate $x_i\in\R^p$ i.i.d.\ from a mean-zero Gaussian
distribution with AR(1) correlation $\Sigma_{ij}=\rho^{|i-j|}$.
Set $(n,p)=(200,400)$. We fix the support set at
$S^\star=\{1,11, \cdots, 181, 191\}$ of size $K=20$, and set
the corresponding nonzero coefficients to
$\{\Phi^{-1}(i/(K+1))\}_{i=1}^K$. We
next generate outcome
\[
  y_i=x_i^\top\betaVec^\star+\varepsilon_i,\qquad
  \varepsilon_i\sim N(0,\sigma^2),
\]\relax
where $\sigma^2$ is chosen so that the signal-to-noise ratio
$\Var(x_i^\top\betaVec^\star)/\sigma^2=3$.
We consider $\rho\in\{0,0.5\}$.

\paragraph{Sparse logistic regression.}
We first generate $x_i$ and $\betaVec^\star$ as above and binary responses from
\[
  \Prob(Y_i=1\mid x_i)
  =
  \bigl\{1+\exp\bigl[-(x_i^\top\betaVec^\star/\sqrt{2})\bigr]\bigr\}^{-1}.
\]\relax

In all settings, we use a 150/50 split for training and validation and an
independent test set of size 20{,}000. Each configuration is repeated for 500 Monte Carlo replicates. 
The evaluation metrics are out-of-sample mean squared error (MSE) for continuous outcomes and AUROC for binary outcomes. We also summarize sparsity by the model size $|\hat S|.$  

\paragraph{Methods compared and tuning.}
\begin{itemize}[leftmargin=*]
  \item \textbf{SAB}: stagewise adaptive boosting with  weight function $p(|\beta|)=(\delta_0+|\beta|^\gamma)^{1/\gamma}-|\beta|$ corresponding to the Bridge penalty, where $\gamma\in\{1,0.8,0.5\}$.
  \item \textbf{Lasso}: the lasso fit with
    \texttt{glmnet} \citep{friedman2010glmnet}.
  \item \textbf{Bridge (LLA)}: a penalized bridge-regression baseline with
    $\gamma=0.5$, fit via local linear approximation
    \citep{zou2008onestep} (iterative reweighted lasso).
\end{itemize}
All three methods are stopped by minimizing validation loss along their
respective paths.

\paragraph{Results.}
We begin with the sparse linear-regression setting.
Table~\ref{tab:sim-linear} summarizes test MSE and the selected model size.
\input{tables/sim_linear_summary.tex}
$\gamma=0.8$ gives the smallest
mean test MSE (7.91) while selecting much smaller models on average (13.5
variables) than either the lasso (8.87 MSE; 62.9 variables) or Bridge LLA (8.19 MSE; 15.0 variables).  
More aggressive concavity ($\gamma=0.5$) yields the sparsest models (10.7
variables) but with a clear predictive cost (8.70 MSE), illustrating the prediction--sparsity trade-off induced by the bridge-functional geometry.
As expected, SAB with $\gamma=1$ (8.84 MSE; 62.0 variables) and the lasso have nearly identical performance, while SAB with $\gamma=0.5$ and Bridge-LLA are likewise quite similar. Under strong correlation ($\rho=0.5$), the lasso and SAB with $\gamma=1$ again remain very close in prediction accuracy (8.75 versus 8.71), whereas SAB with $\gamma=0.8$ yields a much smaller model (14.3 variables) together with improved prediction accuracy (8.19 MSE).  Table~\ref{tab:sim-linear} also reports the corresponding mean test AUC and selected model size, arranged by correlation level, for binary outcomes. SAB with $\gamma=0.8$ selects substantially smaller models ($\sim 40$ vs $6$ predictors), while achieving predictive accuracy comparable to that of logistic lasso regression.  

\subsection{Simulation study for multi-outcome extension}
To examine the proposed multi-outcome setting when outcome types differ, we
also conduct a two-outcome simulation with one continuous and one
binary response generated as in the previous simulations.  Thus the two outcomes models share the same set of predictive features. $\eta_m\in \{1, 4, 7, 10\}$ is selected using the validation set. The reference methods compared are lasso and SAB, each fit separately to the continuous and binary outcome models. This isolates whether modified algorithm benefits from the shared support pattern. As in Section~7.1, we use the same training/validation/test splitting scheme and 500 Monte Carlo replications. Table~\ref{tab:sim-multioutcome-summary} summarizes the results. In this mixed-outcome experiment, modified SAB algorithm  achieves a better prediction performance than separating fitting.  
\input{tables/sim_multioutcome_summary.tex}

\section{Data Applications}\label{sec:data}
We illustrate SAB on two public datasets and focus on the interplay between
predictive accuracy and sparsity. Code to reproduce the analyses is provided
in the \texttt{codes/data} folder, together with scripts that record software
versions to support reproducibility.

We consider the Communities and Crime dataset from the UCI Machine Learning
Repository \citep{redmond2002communities,redmondbaveja2002}, which combines socio-economic
covariates from the 1990 US Census with law-enforcement information and FBI
Uniform Crime Reports.
The response is the per-capita violent-crime rate
(\texttt{ViolentCrimesPerPop}).
After removing identifier variables and the provided fold index, we retain
122 numeric predictors; missing predictor values are imputed by thier corresponding median values from the training data within each split.

To evaluate predictive performance we use the 10-fold split provided with the
dataset as an \emph{outer} evaluation loop.
Within each outer training set we use an additional $80/20$ split to tune the
regularization level by validation loss, and we then refit the tuned model on
the full outer training set before evaluating on the outer test fold.
We compare SAB with $\gamma\in\{1,0.8,0.5\}$ to the lasso and bridge-LLA.

To highlight interaction selection under hierarchical constraints, the
Appendix repeats this analysis after screening to a small set of
main effects and augmenting the design with all pairwise interactions.
We compare lasso and SAB on the expanded design, Ordered-SAB enforcing strong
heredity through prerequisite-closure updates (an interaction can enter only by
paying for any missing main effects), and
\texttt{glinternet} \citep{lim2015glinternet} as a strong-hierarchy baseline.
These interaction-screened results provide an empirical demonstration of the
closure-based prerequisite accounting mechanism from Section~\ref{sec:extensions}
in a real-data setting (see Appendix, ``Communities and Crime with
screened interactions'').
Briefly, in this interaction-screened study, unconstrained fits on the expanded
design can select many interactions without both main effects, whereas
Ordered-SAB and \texttt{glinternet} enforce strong heredity by construction
(zero hierarchy violations) while retaining comparable outer-fold MSE with much
smaller selected models.
The Appendix also reports stability summaries showing that the
screened main effects selected by Ordered-SAB are highly consistent across
folds, and that a small set of interactions recurs across eligible folds,
yielding interpretable candidates for effect modification.
In particular, the interaction \texttt{racepctblack:racePctWhite} is selected
in $7/10$ eligible outer folds, and \texttt{PctPersOwnOccup:racepctblack} in
$4/10$ eligible folds (Appendix, ``Communities and Crime with
screened interactions'').

\input{tables/data_communities_summary.tex}
\begin{figure}[t]
  \centering
  \includegraphics[width=0.95\textwidth]{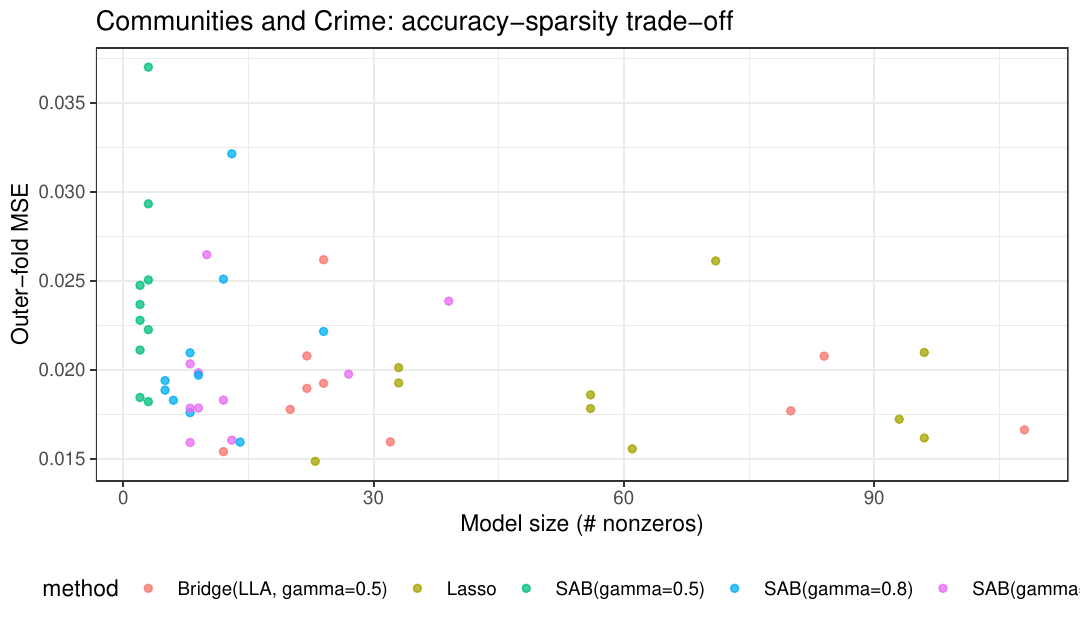}
  \caption{Communities and Crime: outer-fold MSE versus selected model size for
  each method.}
  \label{fig:communities-tradeoff}
\end{figure}
Table~\ref{tab:data-communities} shows that the lasso achieves the smallest
outer-fold MSE but selects a substantially larger model.
In contrast, SAB with $\gamma=1$ attains comparable predictive accuracy while
selecting an order of magnitude fewer variables on average (14 vs. 62 selected predictors).
Figure~\ref{fig:communities-tradeoff} highlights the corresponding
accuracy--sparsity trade-off across outer folds.
The Appendix reports additional stability diagnostics that
summarize which predictors are repeatedly selected across folds.

\section{Discussion}\label{sec:discussion}
We have proposed a stagewise boosting algorithm that operates under an
$\ell_\gamma$-type bridge functional on the coefficient vector.
By parameterizing the path directly in terms of the bridge-functional level and
deriving the corresponding coordinate steps, the method provides a transparent
link between concave regularization and incremental boosting-style updates.
The resulting stagewise adaptive boosting trajectories interpolate between
$\ell_1$-constrained paths and more aggressively sparse concave-penalty paths,
while retaining the computational simplicity of coordinate-wise updates.

Two limitations deserve emphasis.
First, for $0<\gamma<1$ the bridge functional is nonconvex, so we do not claim that
the stagewise path coincides with a globally optimal constrained solution path
at each bridge-functional level.
Second, several statements in Section~\ref{sec:theory} are heuristic and are
intended to clarify the geometry of the updates rather than provide a complete
convergence theory.

Several directions for further research remain. On the theoretical side, a
rigorous analysis of the limiting dynamics behind the local delayed-entry
principle in Section~\ref{sec:theory}, including existence,
uniqueness, switching-time comparisons, and convergence to stationary points
of the constrained problem, would deepen understanding of the method.
On the methodological side, extensions to grouped, structured, and
nonparametric base learners, as well as adaptive tuning of the budget cap and
concavity along the path, appear promising. Broader empirical studies would
help clarify when stagewise adaptive boosting offers substantial advantages
over existing regularization and boosting techniques.

\bibliographystyle{plainnat}
\bibliography{./bib/paper_main}

\appendix

\FloatBarrier
\section{Appendix}
\subsection{Numerical study on global sparsity}
\begin{figure}[t]
  \centering
  \includegraphics[width=0.9\textwidth]{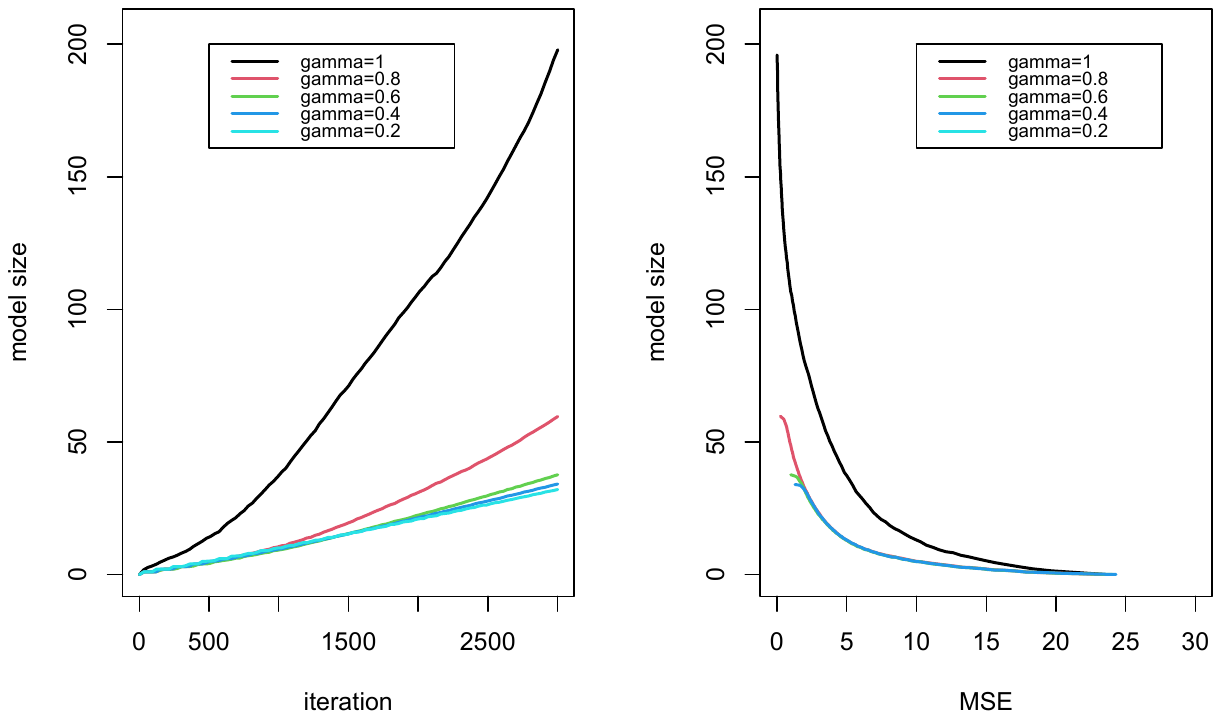}
  \caption{Average model size along the solution paths generated by the SAB algorithm.}
  \label{fig:sparsity}
\end{figure}

We conducted a simulation study to examine the sparsity of the resulting solution paths. To this end, we used the same simulation setup as in Section~4. We generated solution paths using the SAB base algorithm with $p(t)=(|t|+\delta_0)^{1/\gamma}-|t|$ and $\gamma\in\{1, 0.8, 0.6, 0.4, 0.2\}$. Figure~\ref{fig:sparsity} summarizes the average model size along the resulting solution paths by plotting model size against the iteration number and the training MSE. As expected, smaller values of $\gamma$ yield sparser solution paths. When model size is plotted against MSE, several solution paths with $\gamma<1$ exhibit similar sparsity levels, suggesting, a $\gamma$ value close to 0.8 is sufficient to encourage additional sparsity in comparison with standard boosting algorithm with $\gamma=1.$

\subsection{ARCENE classification}
\begin{figure}[t]
  \centering
  \includegraphics[width=0.9\textwidth]{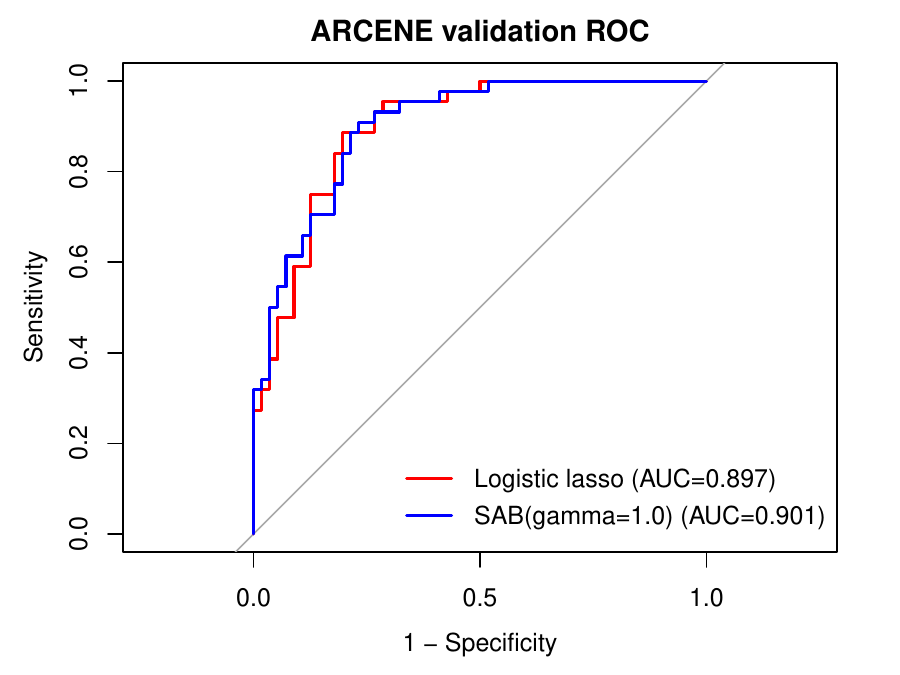}
  \caption{ARCENE: ROC curves for the best-performing SAB fit
  (among the $\gamma$ grid) and the logistic lasso, after screening to 1{,}000
  features in the validation set.}
  \label{fig:arcene-roc}
\end{figure}

We analyze the ARCENE mass-spectrometry dataset from the NIPS 2003 feature
selection challenge \citep{guyon2004arcene} as a second real data example.
The dataset provides a binary classification problem with 200 training
observations, 100 validation observations, and 10{,}000 features (including a
set of ``probe'' variables with no predictive power).
We use the official training/validation split.
To reduce dimension and improve comparability, we apply a simple univariate
screening step on the training set, retaining the 1{,}000 features with the
largest absolute correlation with the training labels; all subsequent fits
use the screened feature set.
Because test labels are not publicly available, we report performance on the
provided validation set.
We compute AUC and ROC curves using the R package \texttt{pROC}
\citep{robin2011proc}. Table~\ref{tab:data-arcene} reports selected model size, and AUC and misclassification error on the validation set.
In this example, SAB and the logistic lasso achieve similar discrimination,
while smaller $\gamma$ values yield markedly sparser classifiers.
Figure~\ref{fig:arcene-roc} compares the corresponding ROC curves.

\begin{table}[ht]
\centering
\begin{tabular}{lllr}
  \toprule
Method & AUC & Error & Model size \\ 
  \midrule
SAB($\gamma=1.0$) & {\color{red} 0.901} & 0.200 &  41 \\ 
  SAB($\gamma=0.8$) & 0.892 & 0.230 &  12 \\ 
  SAB($\gamma=0.5$) & 0.828 & 0.220 &  {\color{red} 6} \\ 
  Logistic lasso & 0.897 & 0.190 &  26 \\ 
   \bottomrule
\end{tabular}
\caption{ARCENE (UCI): validation-set AUC, misclassification error, and selected model size after screening to the top 1000 features.} 
\label{tab:data-arcene}
\end{table}

\input{sections/data_communities_interactions_supp.tex}

\end{document}

%% file: tables/sim_linear_summary.tex
\begin{table}[ht]
\centering
\begin{tabular}{lcccc}
  \toprule
& \multicolumn{2}{c}{$\rho = 0$} & \multicolumn{2}{c}{$\rho = 0.5$} \\
\cmidrule(lr){2-3} \cmidrule(lr){4-5}
&\multicolumn{4}{c}{Continuous outcomes}\\
method & MSE & Model size & MSE & Model size \\
\cline{2-5}\\
Lasso                      & 8.87 (0.05) & 62.9 (1.0) & 8.75 (0.04) & 60.3 (0.9) \\
SAB($\gamma$=1)            & 8.84 (0.04) & 62.0 (1.0) & 8.71 (0.04) & 60.8 (0.9) \\
SAB($\gamma$=0.8)          & {\color{red} 7.91 (0.05)} & 13.5 (0.2) & {\color{red} 8.19 (0.06)} & 14.3 (0.2) \\
SAB($\gamma$=0.5)          & 8.70 (0.07) & {\color{red} 10.7 (0.1)} & 9.19 (0.08) & {\color{red} 10.7 (0.2)} \\
Bridge (LLA, $\gamma$=0.5) &8.19 (0.05) & 15.0 (0.3) & 8.27 (0.05) & 16.5 (0.3) \\
&\multicolumn{4}{c}{Binary outcomes}\\
       & AUC & Model size & AUC & Model size \\
\cline{2-5}\\
Logistic lasso    & 0.76 (0.002) & 39.2 (0.8) & 0.77 (0.002) & 37.9 (0.8) \\
SAB($\gamma$=1)   & {\color{red} 0.77 (0.002)} & 39.3 (0.8) & {\color{red} 0.77 (0.002)} & 38.6 (0.8) \\
SAB($\gamma$=0.8) & 0.76 (0.003) & 6.2  (0.2) & 0.75 (0.003) & 6.2  (0.2) \\
SAB($\gamma$=0.5) & 0.74 (0.003) & {\color{red} 4.0  (0.1)} & 0.73 (0.003) & {\color{red} 4.1  (0.1)}\\
   \bottomrule
\end{tabular}
\caption{Regression simulation under the fixed-support design $S^\star=\{1,11, \ldots, 171, 191\}$ with nonzero coefficients $\{\Phi^{-1}(i/(K+1))\}_{i=1}^K$. Entries report mean test MSE and selected model size for continuous outcomes and mean test AUC and selected model size for binary outcomes, with Monte Carlo standard errors in parentheses. }
\label{tab:sim-linear}
\end{table}

%% file: tables/sim_multioutcome_summary.tex
\begin{table}[ht]
\centering
\begin{tabular}{llcccr}
  \toprule
  & & \multicolumn{2}{c}{Continuous outcome} & \multicolumn{2}{c}{Binary outcome} \\
  \cline{3-4} \cline{5-6}
  $\rho$& Method & MSE & Model size & AUC & Model size \\
  \midrule
 0 &  Lasso & 8.87 (0.05) & 62.9 (1.0) & 0.76 (0.002) & 39.2 (0.8) \\
 0 & SAB~~~~~~ ($\gamma=1.0$) & 8.84 (0.04) & 62.0 (1.0) & 0.77 (0.002) & 39.3 (0.8) \\
 0 & SAB-MO ($\gamma=1.0$) & 7.71 (0.03) & 44.9 (0.7) & {\color{red} 0.81 (0.001)} & 34.9 (0.7) \\
 0 & SAB~~~~~~ ($\gamma=0.8$) & 7.91 (0.05) & 13.5 (0.2) & 0.76 (0.003) & {\color{red} 6.2 (0.2)} \\
 0 & SAB-MO ($\gamma=0.8$) & {\color{red} 7.42 (0.04)} & {\color{red} 13.2 (0.2)} & 0.79 (0.002) & 6.3 (0.1) \\
0.5 & Lasso &8.71 (0.04) & 60.8 (0.9) & 0.77 (0.002) & 38.6 (0.8)\\
0.5 & SAB~~~~~~ ($\gamma=1.0$)   & 8.71 (0.04) & 60.8 (0.9) & 0.77 (0.002) & 38.6 (0.8)\\
0.5 & SAB-MO ($\gamma=1.0$)& 7.72 (0.03) & 44.7 (0.7) & {\color{red} 0.81 (0.001)} & 33.4 (0.7)\\
0.5 & SAB~~~~~~ ($\gamma=0.8$)   & 8.19 (0.06) & 14.3 (0.2) & 0.75 (0.003) & {\color{red} 6.2 (0.2)}\\
0.5 & SAB-MO ($\gamma=0.8$)& {\color{red} 7.59 (0.04)} & {\color{red} 13.6 (0.2)} & 0.78 (0.002) & 8.5 (0.2)\\
  \bottomrule
\end{tabular}
\caption{Simulation results of the mixed continuous--binary two-outcome simulation.}
\label{tab:sim-multioutcome-summary}
\end{table}

%% file: tables/data_communities_summary.tex
\begin{table}[ht]
\centering
\begin{tabular}{lll}
  \toprule
method & CV MSE & Model size \\ 
  \midrule
Lasso & {\color{red} 0.0187 (0.0010)} & 61.8 (8.6) \\ 
  Bridge(LLA, gamma=0.5) & 0.0189 (0.0010) & 42.8 (10.8) \\ 
  SAB(gamma=1.0) & 0.0196 (0.0011) & 14.3 (3.3) \\ 
  SAB(gamma=0.8) & 0.0210 (0.0015) & 10.4 (1.8) \\ 
  SAB(gamma=0.5) & 0.0243 (0.0018) & {\color{red} 2.5 (0.2)} \\ 
   \bottomrule
\end{tabular}
\caption{Communities and Crime (UCI): outer-fold mean squared error (SE) and average selected model size (SE) across the 10 outer folds.} 
\label{tab:data-communities}
\end{table}

%% file: sections/data_communities_interactions_supp.tex

\subsection{Communities and Crime with screened interactions}
\label{sec:supp-communities-interactions}

To illustrate the ordered/hierarchical extension in a real-data setting, we repeat the Communities and Crime analysis from Section~\ref{sec:data} after augmenting a screened predictor set with all pairwise interactions. Within each fold, we screen to select the top $m=20$ main-effect predictors by absolute correlation with the response (computed on the inner training set), form the expanded predictor set consisting of the $20$ main effects and all
$m(m-1)/2=190$ pairwise products, and train outcome prediction models, which are tuned by inner validation set. We compare lasso and SAB with the expanded predictor set, Ordered-SAB enforcing strong
heredity on the expanded predictor set, and \texttt{glinternet}
\citep{lim2015glinternet} as another reference method with strong main effects and interactions hierarchical constraints.

\input{tables/data_communities_interactions_summary.tex}
Table~\ref{tab:data-communities-interactions} shows that Ordered-SAB successfully enforces strong heredity, while achieving MSE from cross-validation comparable to lasso, the best baselines, in this screened setting. With the main-effects and interactions constraint, the Ordered-SAB and glinternet have almost the same predictive performance. To examine the robustness of findings from Ordered-SAB,  Table~\ref{tab:data-communities-interactions-top}
reports the most frequently selected main effects and interactions under Ordered-SAB across 10 outer folds. Because the screened dictionary varies by fold, we report frequencies as selected/eligible folds, where eligibility is determined by whether both main effects parents appear in the screened main-effect set.
\input{tables/data_communities_interactions_top_terms.tex}

In this sensitivity analysis, we repeat the interaction study using screening thresholds
$m\in\{15,25\}$. Table~\ref{tab:data-communities-interactions-screen-sensitivity} shows that the accuracy--sparsity trade-off is stable across this range. As $m$ increases, the candidate interaction dictionary expands and more main effects and interactions are selected. The qualitative comparisons among methods remain similar: lasso achieves the lowest MSE, Ordered-SAB and glinternet have comparable predictive performance, and baseline SAB selects the smallest models.
\input{tables/data_communities_interactions_screen_sensitivity.tex}

%% file: tables/data_communities_interactions_summary.tex
\begin{table}[ht]
\centering
\begin{tabular}{llll}
  \toprule
method & CV MSE & Model size & Violation rate \\ 
  \midrule
Lasso (expanded) & {\color{red} 0.0190 (0.0009)} & 81.7 (14.6) & 0.860 (0.073) \\ 
  glinternet (screened) & 0.0196 (0.0011) & 21.4 (0.9) & {\color{red} 0.000 (0.000)} \\ 
  Ordered-SAB (expanded) & 0.0200 (0.0010) & 23.7 (1.4) & {\color{red} 0.000 (0.000)} \\ 
  SAB (expanded) & 0.0213 (0.0011) & {\color{red} 11.9 (1.5)} & 0.994 (0.006) \\ 
   \bottomrule
\end{tabular}
\caption{Communities and Crime (UCI): interaction analysis. Outer-fold mean squared error (SE), average selected model size (SE), and hierarchy violation rate (SE) across the 10 outer folds.} 
\label{tab:data-communities-interactions}
\end{table}

%% file: tables/data_communities_interactions_top_terms.tex
\begin{table}[ht]
\centering
\begingroup\scriptsize
\begin{tabular}{lclc}
  \toprule
Main effect & Selected/eligible folds & Interaction & Selected/eligible folds (int.) \\ 
  \midrule
MalePctDivorce & 10/10 & racepctblack:racePctWhite & 7/10 \\ 
  PctFam2Par & 10/10 & PctPersOwnOccup:racepctblack & 4/10 \\ 
  PctIlleg & 10/10 & PctHousLess3BR:PctKids2Par & 4/9 \\ 
  PctKids2Par & 10/10 & MalePctDivorce:PctHousLess3BR & 3/9 \\ 
  PctPopUnderPov & 10/10 & FemalePctDiv:PctKids2Par & 2/10 \\ 
  PctYoungKids2Par & 10/10 & FemalePctDiv:PctYoungKids2Par & 2/10 \\ 
   \bottomrule
\end{tabular}
\endgroup
\caption{Communities and Crime (UCI): interaction-screened analysis. Most frequently selected main effects and interactions under Ordered-SAB. Counts are selected/eligible outer folds, where eligibility is determined by the screening step (top m=20 main effects per fold).} 
\label{tab:data-communities-interactions-top}
\end{table}

%% file: tables/data_communities_interactions_screen_sensitivity.tex
\begin{table}[ht]
\centering
\begin{tabular}{rllll}
  \toprule
Screen size m & method & CV MSE & Model size & Violation rate \\ 
  \midrule
 15 & Ordered-SAB (expanded) & 0.0213 (0.0010) & 18.3 (1.5) & 0.000 (0.000) \\ 
   15 & glinternet (screened) & 0.0204 (0.0012) & 15.9 (1.4) & 0.000 (0.000) \\ 
   15 & SAB (expanded) & 0.0212 (0.0010) & 9.0 (1.4) & 1.000 (0.000) \\ 
   15 & Lasso (expanded) & 0.0194 (0.0010) & 63.5 (9.1) & 0.758 (0.075) \\ 
   25 & Ordered-SAB (expanded) & 0.0199 (0.0012) & 27.6 (2.9) & 0.000 (0.000) \\ 
   25 & glinternet (screened) & 0.0191 (0.0011) & 26.0 (1.2) & 0.000 (0.000) \\ 
   25 & SAB (expanded) & 0.0203 (0.0009) & 11.5 (0.9) & 0.993 (0.007) \\ 
   25 & Lasso (expanded) & 0.0185 (0.0010) & 78.3 (15.2) & 0.937 (0.027) \\ 
   \bottomrule
\end{tabular}
\caption{Communities and Crime (UCI): robustness to the screening threshold $m$ in the interaction-screened analysis. For each $m$, we screen to the top-$m$ main effects within each outer fold, expand to all pairwise interactions among screened predictors, tune by inner validation, and report mean (Monte Carlo SE) of outer-fold MSE, selected model size, and hierarchy violation rate.} 
\label{tab:data-communities-interactions-screen-sensitivity}
\end{table}